\documentclass[runningheads]{llncs}

\usepackage[mobile]{eccv}

\usepackage{eccvabbrv}

\usepackage{graphicx}
\usepackage{booktabs}
\usepackage{wrapfig}
\usepackage{floatflt}

\usepackage[accsupp]{axessibility}  
\usepackage[table]{xcolor}
\usepackage{soul}
\usepackage{color, xcolor}

\usepackage{hyperref}

\usepackage{orcidlink}

\newcommand{\parahead}[1]{\vspace{1mm}\noindent\textbf{#1.}\ }
\newcommand{\ours}{MoCam\xspace}

\begin{document}

\title{MoCam: Unified Novel View Synthesis via Structured Denoising Dynamics} 

\titlerunning{MoCam}

\author{Haofeng Liu \and
Yang Zhou \and
Ziheng Wang \and
Zhengbo Xu \and
Zhan Peng \and
Jie Ma \and
Jun Liang\and
Shengfeng He \and
Jing Li}

\authorrunning{Liu et al.}

\institute{
Project Page: \url{https://orange-3dv-team.github.io/MoCam/}
}

\maketitle

\begin{figure}[h]
    \centering
    \centering
    \vspace{-7mm}\includegraphics[width=\linewidth]{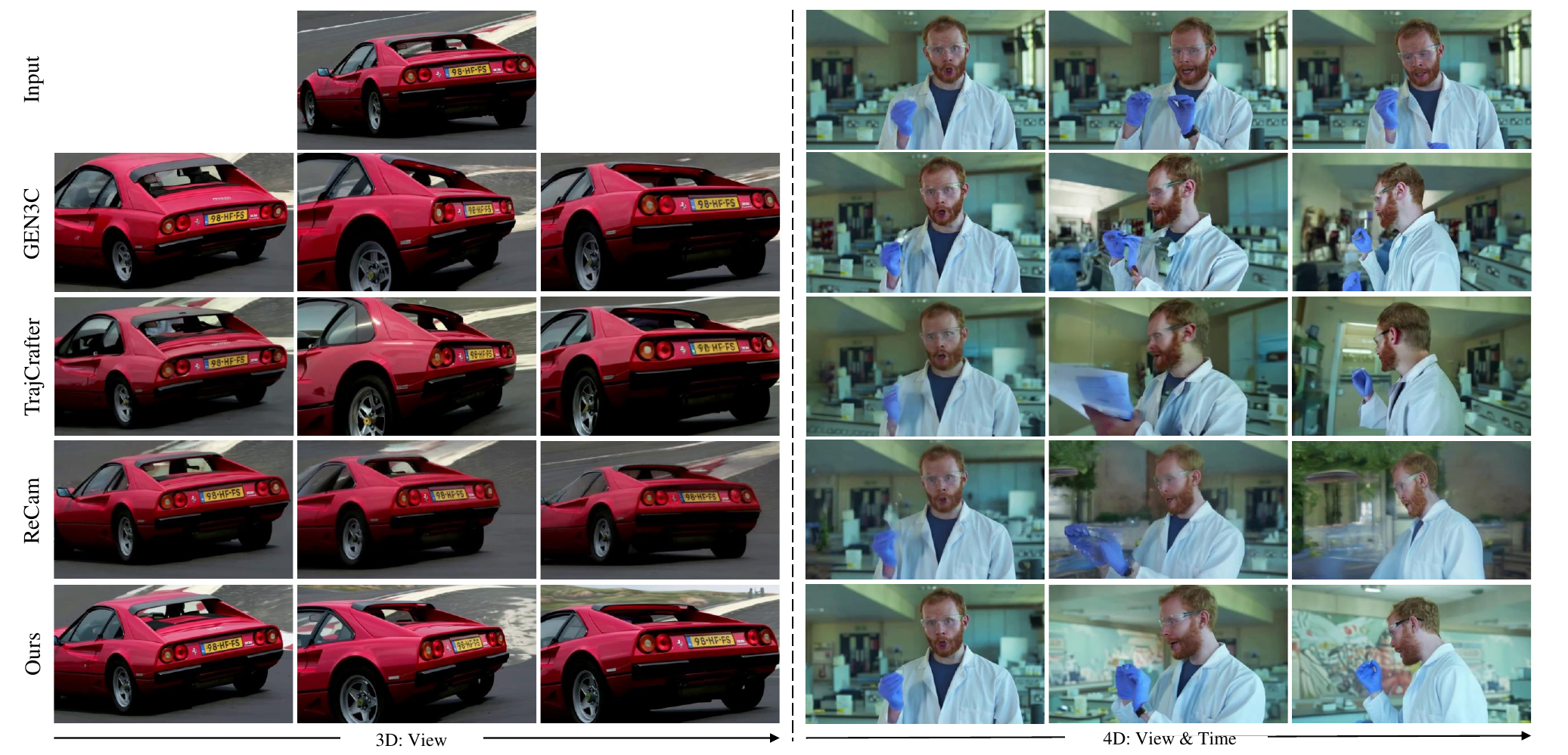}
    \vspace{-6mm}
    \caption{We propose MoCam, a method that unifies novel view synthesis through structured denoising dynamics. Existing methods rely on static guidance that entangles geometry and appearance, often resulting in geometric collapse and visual artifacts. MoCam introduces structured denoising dynamics that guide generation from motion alignment to appearance refinement, producing coherent and photorealistic results.
    }\vspace{-7mm}
    \label{fig:teaser}
\end{figure}

\begin{abstract}
    Generative novel view synthesis faces a fundamental dilemma: geometric priors provide spatial alignment but become sparse and inaccurate under view changes, while appearance priors offer visual fidelity but lack geometric correspondence. Existing methods either propagate geometric errors throughout generation or suffer from signal conflicts when fusing both statically. 
    We introduce MoCam, which employs structured denoising dynamics to orchestrate a coordinated progression from geometry to appearance within the diffusion process.
    MoCam first leverages geometric priors in early stages to anchor coarse structures and tolerate their incompleteness, then switches to appearance priors in later stages to actively correct geometric errors and refine details. 
    This design naturally unifies static and dynamic view synthesis by temporally decoupling geometric alignment and appearance refinement within the diffusion process.
    Experiments demonstrate that MoCam significantly outperforms prior methods, particularly when point clouds contain severe holes or distortions, achieving robust geometry-appearance disentanglement.
\end{abstract}

\section{Introduction}
\label{sec:intro}

Novel view synthesis aims to create photorealistic views from arbitrary camera trajectories given limited input, and it remains a fundamental challenge in computer vision with broad applications in virtual production, immersive reality, and content creation. This encompasses two closely related problems: single-image 3D reconstruction, where a static scene is reconstructed from one photograph, and video 4D re-camera, where dynamic scenes are rendered along new camera paths given a monocular video. Success in both settings requires reconciling precise geometric control with high-fidelity appearance synthesis, particularly when the target viewpoint significantly deviates from the input.

Recent advances in diffusion models~\cite{blattmann2023stable, wan2025, yang2024cogvideox, kong2024hunyuanvideo} have enabled impressive progress in both domains. For 3D reconstruction, methods like ViewCrafter~\cite{yu2025viewcrafter}, VistaDream~\cite{wang2025vistadream}, and SpatialCrafter~\cite{zhang2025spatialcrafter} leverage reconstructed geometry to guide novel view synthesis. For 4D re-camera, approaches such as Gen3C~\cite{ren2025gen3c} and TrajectoryCrafter~\cite{yu2025trajectorycrafter} employ 3D scaffolds (\eg, point clouds) to render target-view videos with explicit camera control. However, these methods share a critical vulnerability: they rely on geometric priors (depth maps or point clouds reconstructed from monocular input) that inevitably become sparse, incomplete, and erroneous under large view changes. Existing pipelines either propagate these geometric flaws throughout generation~\cite{ren2025gen3c, yu2025trajectorycrafter} or attempt to fuse geometry and appearance statically, causing signal conflicts that degrade both structure and texture (see Fig.~\ref{fig:teaser}), limiting their applicability in settings that demand high-fidelity and precise cinematic control.

We argue that this bottleneck stems from a fundamental tension between two complementary yet incompatible signal sources. Rendered geometric scaffolds provide essential spatial alignment with target trajectories but suffer from holes and distortions due to disocclusion and depth inaccuracy. Conversely, source images/videos offer rich, high-fidelity appearance but are geometrically misaligned with novel views. Crucially, these signals cannot be effectively combined simultaneously: early in generation, strong appearance cues dominate and cause geometric drift; late in generation, flawed geometry permanently bakes structural errors into the output.

To resolve this, we introduce \ours, a framework that exploits \textit{structured denoising dynamics} to temporally decouple geometry and appearance priors within the diffusion process. Our key insight is that diffusion models exhibit distinct representational needs across denoising phases: early stages require coarse structural anchoring, while later stages demand high-frequency refinement. \ours orchestrates a coordinated progression: in early timesteps, the model conditions solely on rendered scaffolds to establish global structure and motion coherence, deliberately tolerating geometric incompleteness. As denoising progresses and the latent stabilizes, \ours transitions to conditioning on the source appearance. At this stage, the established geometry enables the model to use appearance not merely for texture transfer, but to actively correct geometric errors and fill disoccluded regions without destabilizing the overall structure.

Notably, this mechanism naturally provides a unified solution for both static and dynamic view synthesis. By structuring the denoising process to first establish geometry and then refine appearance, MoCam separates geometric alignment from appearance synthesis in a manner that is independent of the input modality. As a result, the same generation principle applies to both single-image 3D view synthesis and video 4D re-camera, highlighting that our approach addresses the underlying challenge of synthesizing views under unreliable geometry.

By transforming denoising into a structured progression from alignment to realism, \ours achieves robust geometry-appearance disentanglement. Even when point clouds contain severe holes or distortions, our method generates geometrically coherent and photorealistic results, significantly outperforming static conditioning approaches (Fig.~\ref{fig:teaser}).

In summary, our contributions are threefold:
\begin{itemize}
    \item We identify the fundamental conflict between geometric and appearance priors in generative view synthesis, and propose \textit{structured denoising dynamics} as a principled solution that temporally decouples these signals.
    \item We present a unified framework for both single-image 3D reconstruction and video 4D re-camera, demonstrating that stage-wise conditioning generalizes across input modalities.
    \item We show that active geometric error correction through late-stage appearance signal achieves state-of-the-art robustness under sparse and inaccurate geometry, setting new standards for controllable view synthesis.
\end{itemize}

\section{Related Works}
\label{sec:related}

\parahead{Optimization-Based Novel View Synthesis}  
A classical approach to novel view synthesis involves reconstructing a 3D or 4D representation from posed images. Neural Radiance Fields (NeRF)~\cite{mildenhall2020nerf} transformed this field by representing a static scene as a continuous volumetric function, enabling unprecedented photorealism. More recently, 3D Gaussian Splatting (3DGS)~\cite{kerbl20233d} has achieved comparable or superior quality with real-time rendering by modeling the scene as a set of explicit 3D Gaussians.
Extending such methods to dynamic scenes, which is essential for video re-camera, requires modeling temporal evolution~\cite{zhu2025dynamic}. One strategy learns 4D representations that map spacetime coordinates to scene properties~\cite{li2021neural, gao2021dynamic, fridovich2023k, cao2023hexplane, yang2023real, li2024spacetime, duan20244d, luo2025instant4d, Zhang_2025_ICCV}, while another explicitly models motion through deformation fields~\cite{pumarola2020d, li2022neural, lin2024gaussian, wu20244d, yang2024deformable, liu2025modgs, Fan_2025_CVPR, Song_2025_ICCV}. Although powerful, these approaches typically require dense multi-view video and involve costly per-scene optimization. When limited to monocular input, both reconstruction quality and appearance fidelity degrade significantly. In contrast, our method avoids per-scene optimization entirely, instead leveraging the generative priors of large-scale video models to synthesize photorealistic and geometrically consistent results from a single input video.

\parahead{Generative Novel View Synthesis}
Recent single-view 3D reconstruction methods~\cite{zhang2024text2nerf, shriram2025realmdreamer, chung2025luciddreamer} leverage pretrained image diffusion models to enable view synthesis from single images. However, generating smooth camera trajectories rather than isolated views requires temporal consistency, motivating the shift to video generative models~\cite{blattmann2023stable, wan2025, yang2024cogvideox, kong2024hunyuanvideo}. For instance, ViewCrafter~\cite{yu2025viewcrafter} harnesses video diffusion to synthesize high-fidelity view sequences along camera paths. 
Extending these methods to dynamic scenes for 4D video re-camera introduces further complexity, as the generation process must simultaneously handle temporal dynamics and viewpoint changes.
Existing approaches fall into two categories. The first injects camera pose information directly into the model’s conditioning mechanism~\cite{bahmani2024vd3d, van2024generative, bai2025recammaster, lei2025motionflow, wu2025cat4d}, offering end-to-end generation but often lacking geometric accuracy, especially for complex or large-scale trajectories. The second category follows a render-then-inpaint strategy~\cite{you2024nvs, zhang2025recapture, jeong2025reangle, ren2025gen3c, yu2025trajectorycrafter, chen2025cognvs}, where a 3D scaffold (e.g., a point cloud) is reconstructed from the source video, rendered along the target path, and then refined using a video inpainting model. Gen3C~\cite{ren2025gen3c} constructs a spatiotemporal 3D cache to guide generation, while TrajectoryCrafter~\cite{yu2025trajectorycrafter} introduces a Ref-DiT block for reference-based conditioning. Although these methods better enforce target-view geometry, they suffer from a key bottleneck: the rendered scaffold is built on sparse and inaccurate geometry, which permanently bakes errors into the generation process. The inpainting stage inherits these flaws and lacks the capacity to correct them. Our method addresses this limitation by introducing a temporally structured guidance strategy. By decoupling geometry and appearance over the denoising process, it mitigates error propagation and improves stability under large camera motions.

\parahead{Conditioning Mechanisms in Diffusion Models}  
Conditioning is the core mechanism for controllability in diffusion models. Techniques such as ControlNet~\cite{zhang2023adding} and T2I-Adapter~\cite{mou2024t2i} allow spatial control using depth maps or other signals, while IP-Adapter~\cite{ye2023ip} enables lightweight image-prompt conditioning. These approaches typically apply static guidance, using the same control signal across all timesteps. 
More recent work has begun to explore dynamic conditioning. TSM~\cite{zhuang2025timestep} and DMP~\cite{ham2025diffusion} demonstrate that adjusting or switching control inputs over time can significantly improve generation quality. Building on this idea, our method introduces a dynamic conditioning scheme tailored to video re-camera. We design a stage-wise handover between two complementary but conflicting inputs: a geometrically aligned yet flawed scaffold, and a view-disaligned but visually rich reference video. This design specifically resolves the error propagation problem by aligning each guidance signal with the stage of denoising where it is most effective.
\begin{figure}[t]
    \centering
    \includegraphics[width=\linewidth]{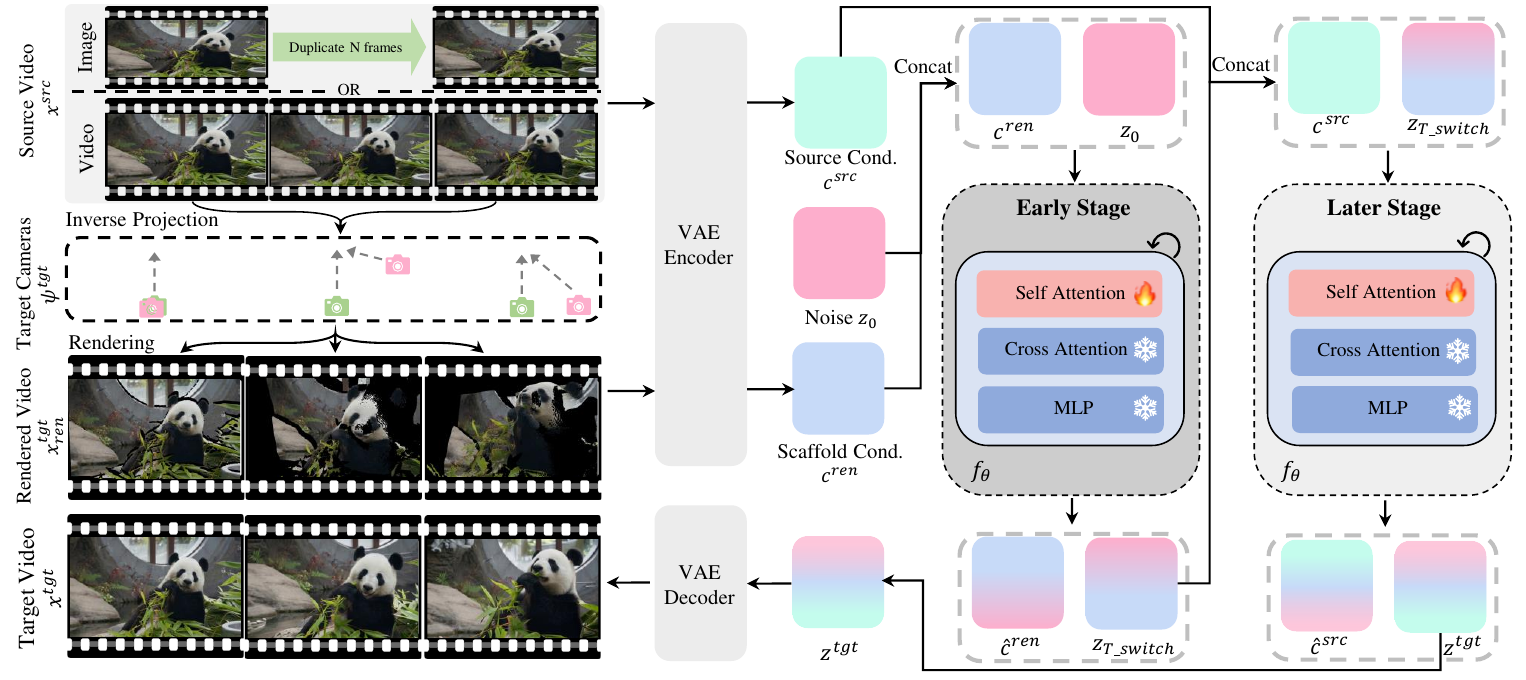}
    \vspace{-7mm}
    \caption{\textbf{Overview of the \ours Framework.} 
    Given a source video $x^{\text{src}}$ (or a single image repeated for N frames), a geometrically-aligned but imperfect scaffold video $x^{\text{tgt}}_{\text{ren}}$ is first rendered along the target trajectory $\psi^{\text{tgt}}$. 
    After encoding these conditions into latent space, the model processes the initial noise $z_0$ via the proposed structured denoising dynamics. Specifically, in the early stage, the denosing is guided by the scaffold condition $c^{\text{ren}}$ to establish geometrically-aligned noise $z_{T\_switch}$. 
    Subsequently, the signal switches to the original source condition $c^{\text{src}}$ to obtain the clean latent $z^{tgt}$ with refined appearance and decode it as the target video $x^{tgt}$.
    This temporal decoupling of conditions prevents the propagation of scaffold errors, enabling stable and photorealistic synthesis.
    }
    \vspace{-8mm}
    \label{fig:pipeline}
\end{figure}

\section{MoCam}
\label{sec:method}

In this section, we present~\ours, a novel framework to generate novel views based on video model.  The core challenge is maintaining geometric and temporal consistency, especially under complex camera movements. Our approach is built on the key insight that different types of conditions are optimal at different stages of the generation process.
Specifically, our method consists of three main stages, as illustrated in Fig.~\ref{fig:pipeline}: (1) We first construct a dynamic point cloud from the monocular input video (or a single image replicated to N frames, \ie, a stationary video) and render it along the target trajectory to create a coarse scaffold video. (2) We then use this scaffold video and the original source video as dual conditioning inputs to our novel stage-Wise generation model. (3) The model first enforces the coarse structure using the scaffold, then switches to the source video to perfect the appearance and geometry.

\subsection{Preliminary: Video Generative Models}
Since our method builds upon a video generative model, we first provide a brief overview of its fundamental principles. For computational efficiency, modern video generative models~\cite{blattmann2023stable, wan2025, yang2024cogvideox, kong2024hunyuanvideo} operate not in the high-dimensional pixel space but in a compressed latent space. This space is constructed by a pre-trained Variational Autoencoders (VAEs)~\cite{wu2025improved}. The VAE consists of an encoder $\mathcal{E}$ that compresses an input video $x \in \mathbb{R}^{N\times H\times W\times 3}$ into a compact latent representation $z=\mathcal{E}(x) \in \mathbb{R}^{n\times h\times w\times c}$, and a decoder $\mathcal{D}$ that reconstructs the video $\hat{x}=\mathcal{D}(z)$ from this latent representation.
Upon this latent space, a generative model $f_\theta$ is trained to model the data distribution. This is typically achieved through one of two primary training paradigms: a denoising diffusion objective or a flow matching objective.
Under the denoising diffusion schema, the model learns to reverse a process that gradually adds noise to the data. The objective is to predict the noise added to a latent representation:
\begin{equation}
   \min_{f_\theta} \mathbb{E}_{z_0, z_1, t, c}\|f_\theta(z_t, t, c) - z_0\|^2_2.
\end{equation}
Alternatively, under the flow matching schema, the model learns a vector field that transports samples from a simple prior distribution to the data distribution:
\begin{equation}
   \min_{f_\theta} \mathbb{E}_{z_0, z_1, t, c}\|f_\theta(z_t, t, c) - v_t\|^2_2,
\end{equation}
where $z_1 = \mathcal{E}(x)$ is the latent encoding of a real video sampled from the data distribution $p_{data}$, and $z_0 \sim \mathcal{N}(0,\mathbf{I})$ is a random latent sampled from a standard Gaussian prior. The variable $t \in [0,1]$ is a continuous time step, and $c$ represents optional conditioning information (such as text prompts or image frames). For the denoising objective, $z_t$ is a noisy latent created by interpolating between $z_1$ and $z_0$ according to a noise schedule (\eg, $z_t = \alpha_t z_1 + \sigma_t z_0$). For the flow matching objective, $z_t$ is typically a linear interpolation $z_t = (1-t)z_0 + t z_1$, and the target velocity is $v_t = z_1 - z_0$.
Crucially, the timestep $t$ represents not merely a noise level, but a progression from global structure to local detail—a property we exploit in our stage-wise conditioning strategy.

\subsection{Scaffold Generation}
The first step of our pipeline is to generate a coarse video draft that is spatially and temporally aligned with the target camera trajectory. This \textit{scaffold video}, denoted as $x^{tgt}_{ren}$, serves as the initial structural guide for our diffusion model.

Given a source video $x^{src} = {\{I^{src}_{i}\}}^N_{i=1} \in \mathbb{R}^{N\times H\times W\times 3}$, we first leverage a depth estimator to acquire its depth $d^{src} = \{D^{src}_i\}^N_{i=1}$. We then follow the inverse perspective projection $\Phi^{-1}$ to construct a dynamic point cloud $p = {\{P_{i}\}}^N_{i=1}$:
\begin{equation}
    p = \Phi^{-1}(x^{src}, d^{src}, K),
\end{equation}
where $K \in \mathbb{R}^{3\times3}$ denotes the camera intrinsic.
We refer this dynamic point cloud $p$ as the 3D scaffold, which provide us a way to precisely control the camera trajectory. Specifically, conditioned by a target camera trajectory $\psi^{tgt} = \{\Psi^{tgt}_i\}^N_{i=1}$, we render the target video $x^{tgt} = {\{I^{tgt}_{i}\}}^N_{i=1} \in \mathbb{R}^{N\times H\times W\times 3}$ from $p$ following the perspective projection $\Phi$:
\begin{equation}
    x^{tgt}_{ren} = \Phi(p, \psi^{tgt}, K).
\end{equation}
As shown in Fig.~\ref{fig:pipeline}, the rendered scaffold video spatially aligns with the target camera motion.
However, due to the inherent limitations of monocular input, this video suffers from significant artifacts: holes from disocclusion, and geometric distortions, particularly in views far from the original camera path. While unsuitable as a final output, it provides an invaluable, spatially-aligned motion prior for the initial stages of generation.

\subsection{Stage-Wise Dual-Conditioning Diffusion}
The proposed latent video generative model integrates conditions from two distinct sources---the scaffold video $x^{tgt}_{ren}$ and the source video $x^{src}$---at different phases of the generation process. We build upon a pretrained latent video diffusion architecture~\cite{wan2025}, which is trained to denoise a noisy latent variable $z_t$ at timestep $t$. \textbf{Our innovation lies in how we formulate the conditioning term $c$.}

We design a stage-wise dual-conditioning architecture. Each stage is responsible for processing one of our condition signals:

\parahead{Spatial Scaffold Condition}
To inject the strong motion and structural prior from the scaffold video $x^{tgt}_{ren}$, we follow the frame dimension conditioning to retain temporal synchronization~\cite{bai2025recammaster}.
Particularly, $x^{tgt}_{ren}$ is first projected into the latent space by the VAE encoder $\mathcal{E}$, $z^{tgt}_{ren} = \mathcal{E}(x^{tgt}_{ren})$,~\ie, conditioning term $c^{ren}$.
After that we concatenate the $c^{ren}$ with the initial noise $z_0$ along the frame dimension as the input of the video model.
This provides direct, spatially-explicit guidance, forcing the generated output to conform to the layout and motion defined by the scaffold.

\parahead{Reference Appearance Condition}
Unlike the scaffold video $x^{tgt}_{ren}$ that contain spatial-aligned information, the source video $x^{src}$ emphasizes high-fidelity appearance and object dynamics of the scene.
It forms a complement relationship with the $x^{tgt}_{ren}$ during the video generation process, in which $x^{tgt}_{ren}$ provide geometry signal and $x^{src}$ supplement the appearance signal.
The conditioning of $x^{src}$ is the same as $x^{tgt}_{ren}$: $\mathcal{E}$, $z^{src} = \mathcal{E}(x^{src})$,~\ie, conditioning term $c^{src}$, then concatenated along the frame dimension.
This mechanism is effective at transferring content and texture, making it ideal for our view-disaligned source video.

\parahead{Why Stage-wise Conditioning is Necessary} 
An intuitive way to leverage these two kinds of condition (\ie, $c^{ren}$ and $c^{src}$) is to concatenate them together with the initial noise $z_0$ and let the model to learn the combinative condition by
\begin{wrapfigure}{r}{6.5cm}
    \vspace{-8mm}
    \centering
    \includegraphics[width=6.5cm]{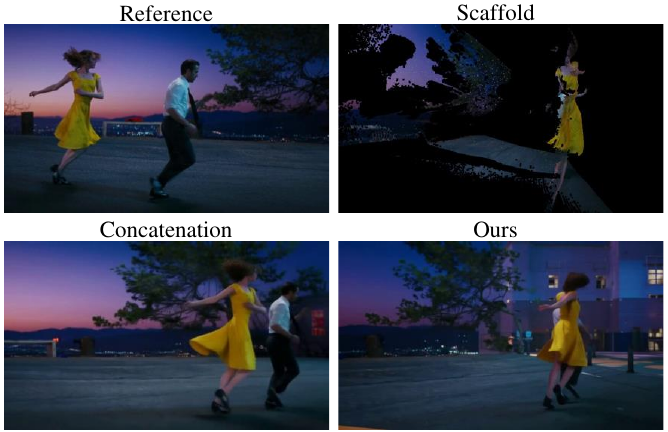}
    \vspace{-6mm}
    \caption{\footnotesize{Results of different guiding methods.}}
    \label{fig:conditioning}
    \vspace{-8mm}
\end{wrapfigure}
itself. Though $c^{ren}$ and $c^{src}$ exhibit the mentioned-above complement relationship, they also contain conflicting signal to each other,~\ie, the different camera movements. Since the camera movement of $c^{src}$ is different with $c^{ren}$, it introduces interference against the guidance of $c^{ren}$, which may confuse the model learning and decrease the final effectiveness. Fig.~\ref{fig:conditioning} illustrates the results of such conditioning. Besides, persistent exposure to $c^{ren}$'s geometric errors causes irreversible structural artifacts. See experiment (Sec.~\ref{sec:ablation}) for detail discussion.

To circumvent this conflict, our approach \ours is motivated by the inherent behavior of diffusion models. We align our conditioning strategy with the progressive denoising process, prioritizing the establishment of global structure in early stages before refining high-frequency details in later ones.
The central novelty of \ours is the \textit{structured denoising dynamics}, where we temporally align these two conditioning signals with respective denoising timesteps $t$. The model's prediction is conditioned on a time-dependent context $c(t)$:
\[
f_\theta(z_t, t, c(t))
\]
where $c(t)$ is defined by a switch at a pre-defined timestep threshold $T_{\text{switch}}$:
\begin{equation}
    c(t) = 
        \begin{cases} 
        c^{ren} & \text{if } t > T_{switch} \\
        c^{src} & \text{if } t \le T_{switch} 
        \end{cases}
\end{equation}

The intuition is as follows:
\begin{itemize}
    \item Early Stage ($t > T_{switch}$): Geometry Anchoring. The latent $z_t$ is mostly noise. The model's primary task is to establish the global structure and motion of the video. By using $c^{ren}$, we force the generation to adhere to the target camera trajectory from the very beginning.
    \item Later Stage ($t \le T_{switch}$): Active Error Correction \& Refinement. The latent $z_t$ already contains a coherent, low-frequency structure that aligns with the target structure. The task now shifts to synthesizing high-frequency details, refining appearance, and correcting geometric inaccuracies. We switch to $c^{src}$, which provides a rich source of clean textures and consistent object appearance. Because the coarse structure is already established, the model can use this high-fidelity reference to ``inpaint'' and ``correct'' the structure inherited from the first stage, without being corrupted by the scaffold's persistent errors.
\end{itemize}
This deliberate handover prevents the scaffold's flaws from being ``baked in'' during the final, high-fidelity synthesis steps, effectively resolving the core limitation of static pipelines.
As shown in Fig.~\ref{fig:pipeline}, the clean latent $z^{tgt}$ is put into the decoder for the final output video $x^{tgt}$:
\begin{equation}
    x^{tgt} = \mathcal{D}(z^{tgt})
\end{equation}

\section{Experiments}
\label{sec:exp}

We implement \ours by building upon the pretrained Wan2.2 video diffusion model~\cite{wan2025} and train it using 20,000 data pairs from the MultiCamVideo dataset~\cite{bai2025recammaster}.
Each training sample consists of a reference video, the resulting scaffold video, and the ground-truth target video.
For scaffold generation, we use ViPE~\cite{huang2025vipe} for depth and camera estimation.
The model is trained for 20,000 steps on eight GPUs with a learning rate of 1e-5 and batch size of 8. $T_{switch}$ is set as 0.85 empirically.

\subsection{Evaluation on In-the-wild Benchmark}
To provide a broad quantitative assessment, we collected 100 monocular videos from OpenVid-1M~\cite{nan2024openvid} and generated outputs for 9 distinct camera trajectories per video, including orbital, translational, and zoom motions. These monocular videos serve as direct input for the 4D re-camera experiments. For single-view 3D reconstruction, we randomly sample one frame from each video and replicate it to N frames.
Our evaluation metrics include: (1) background consistency, subject consistency and imaging quality from VBench metrics~\cite{huang2023vbench}, (2) FVD-V and CLIP-V that calculate FVD and CLIP scores between different viewpoints, (3) pose accuracy: rotation error and translation error~\cite{he2024cameractrl}. 

\begin{table}[t]
	\caption{Quantitative 3D reconstruction comparisons on the OpenVid dataset. BC: Background Consistency, SC: Subject Consistency, IQ: Imaging Quality, RotErr: Rotation Error, TransErr: Translation Error. Cells highlighted in \sethlcolor{red!18}\hl{red} and \sethlcolor{yellow!22}\hl{yellow} denote the best and second-best performance.}
    \vspace{-3mm}
	\label{tab:quantitative_3d}
    \centering
    \footnotesize
    \setlength\tabcolsep{0.5mm}
    \begin{tabular}{c|ccc|cc|cc}
        \toprule
        \hline
        & \multicolumn{3}{c|}{VBench} & \multicolumn{2}{c|}{Perceptual Quality} & \multicolumn{2}{c}{Pose Accuracy} \\
		\hline      
		Methods & BC $\uparrow$ & SC $\uparrow$ & IQ $\uparrow$ & FVD-V $\downarrow$ & CLIP-V $\uparrow$ & RotErr $\downarrow$ & TransErr $\downarrow$\\
        \hline
		GEN3C & \cellcolor{yellow!22}0.9295 & \cellcolor{yellow!22}0.9076 & \cellcolor{yellow!22}0.6909 &  \cellcolor{yellow!22}289.37 &  \cellcolor{yellow!22}0.80 & \cellcolor{yellow!22} 1.36 &  \cellcolor{yellow!22}5.12\\
        \hline
		TrajCrafter &0.9241 & 0.9070 & 0.6700 &  313.65 &  0.79 & \cellcolor{yellow!22}1.36 &  \cellcolor{red!18}5.11 \\
        \hline
		ReCam & 0.9016 & 0.8818 & 0.5842 &  355.26 &  0.77 &  2.13 &  5.79 \\
        \hline
        \hline
        Ours  & \cellcolor{red!18}0.9334 & \cellcolor{red!18}0.9250 & \cellcolor{red!18}0.6961 & \cellcolor{red!18} 255.16 &  \cellcolor{red!18}0.87 &  \cellcolor{red!18}1.35 &  \cellcolor{red!18}5.11 \\
        \hline
    \bottomrule
	\end{tabular}
    \vspace{-4mm}
\end{table}

\begin{figure}[t]
  \centering
  \includegraphics[trim={0 0 0 0},clip,width=\linewidth]{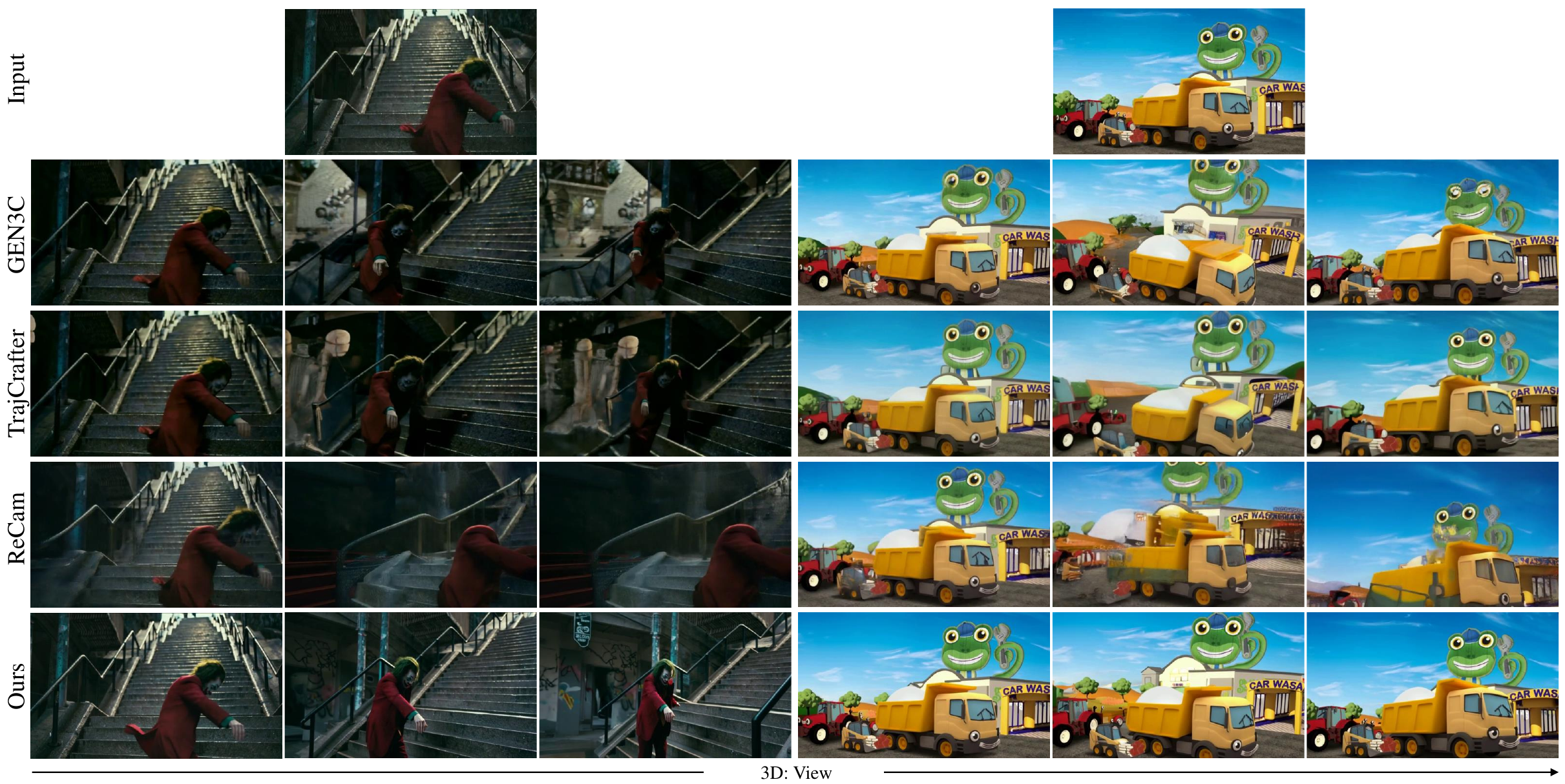}
  \vspace{-8mm}
  \caption{\footnotesize Qualitative results for single-view 3D reconstruction.}
  \vspace{-8mm}
  \label{fig:qualitative_3d}
\end{figure}

\begin{figure}[t]
  \centering
  \vspace{-2mm}\includegraphics[trim={0 0 0 0},clip,width=\linewidth]{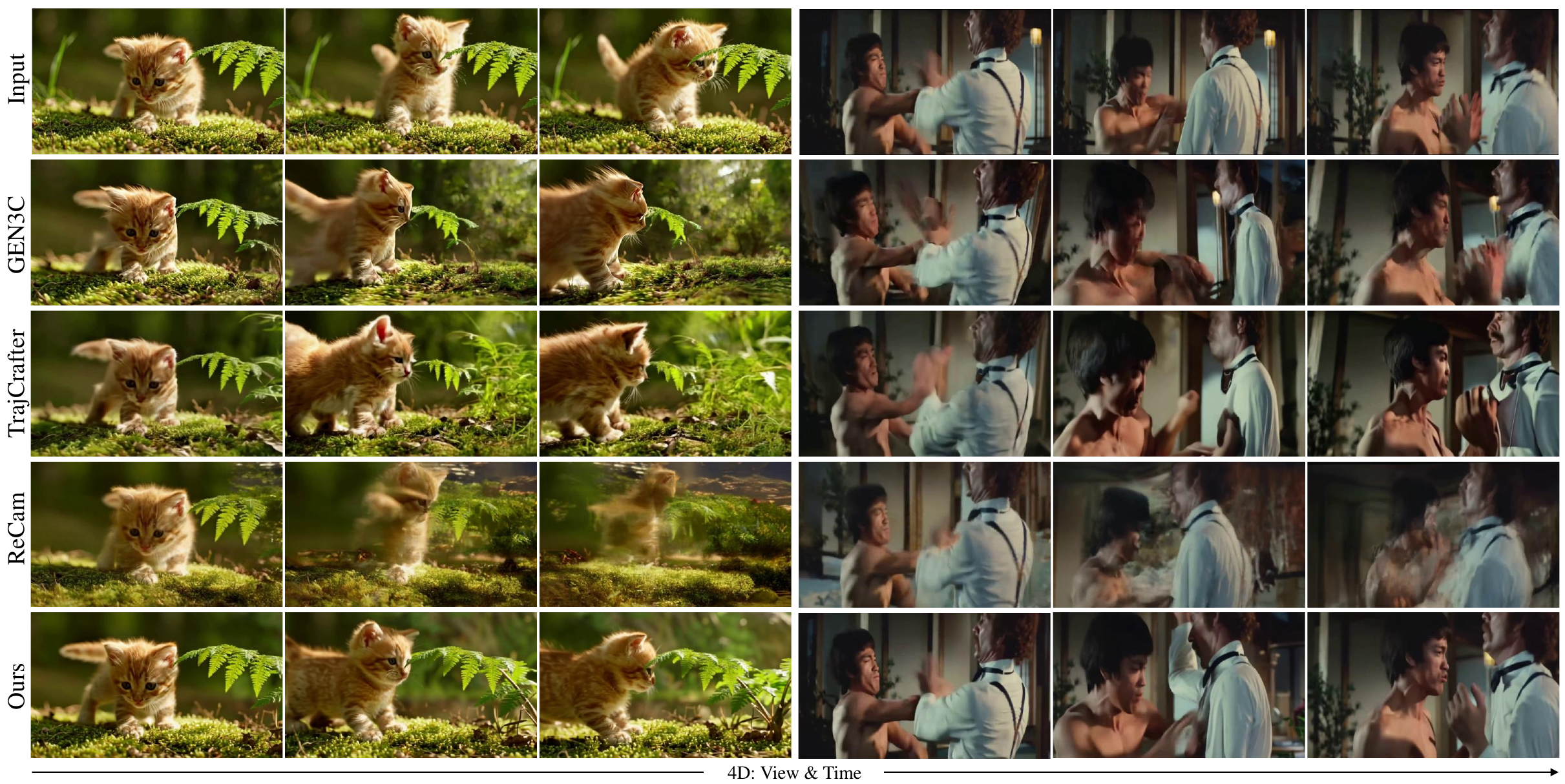}
  \vspace{-8mm}
  \caption{\footnotesize Qualitative results from in-the-wild videos. The first example illustrates an 'orbit-to-left' trajectory, while the second example demonstrates a camera motion that initially moves to the top-left with zoom-in, followed by a transition to the bottom-right with a corresponding zoom-out.}
  \vspace{-6mm}
  \label{fig:in-the-wild}
\end{figure}

\parahead{3D Reconstruction Qualitative Results}
Fig.~\ref{fig:qualitative_3d} visualizes single-view synthesis results. GEN3C and TrajCrafter struggle with the extreme sparsity of single-image point clouds, leading to structural distortions. ReCamMaster fails to infer correct 3D layouts without explicit geometry. In contrast, \ours leverages structured denoising dynamics to overcome this sparsity: we first anchor plausible geometry using the limited scaffold, then refine appearance, yielding coherent and detailed results.

\parahead{3D Reconstruction Quantitative Results}
Tab.~\ref{tab:quantitative_3d} demonstrates our superiority. \ours significantly outperforms competitors in perceptual quality (e.g., FVD-V 255.16 vs. 289.37) and achieves the lowest pose errors. This confirms that our structured denoising strategy effectively maintains both high perceptual fidelity and precise camera control, effectively handling the geometric ambiguity inherent in single-view 3D reconstruction.

\parahead{4D Re-Camera Qualitative Results}
Fig.~\ref{fig:in-the-wild} presents a qualitative comparison against state-of-the-art methods across diverse scenes and camera trajectories. Methods based on 3D scaffolds, like GEN3C~\cite{ren2025gen3c} and TrajectoryCrafter~\cite{yu2025trajectorycrafter}, successfully follow the target movement but suffer from severe geometric degradation. For instance, in the first example, the cat's body becomes distorted as the camera orbits. Similarly, in the second example, the human's arm collapses unrealistically. These artifacts are direct consequences of error propagation from the sparse and inaccurate point cloud reconstruction. The implicit conditioning method, ReCamMaster~\cite{bai2025recammaster}, struggles to maintain geometric consistency and fails to follow the complex trajectory, resulting in chaotic and unusable outputs. In contrast, \ours generates results that are both geometrically coherent and photorealistic. Our method correctly preserves the 3D structure of the subjects (the cat's volume, the person's limbs) while rendering high-fidelity textures, even under significant view changes. 
We provide more dynamic results in the supplementary video.

\parahead{4D Re-Camera Quantitative Results}
As shown in Tab.~\ref{tab:vbench}, \ours achieves the highest scores across the majority of metrics, notably in background consistency, subject consistency, and imaging quality, outperforming all competitors by significant margins. This confirms that our method not only maintains the identity and structure of the main subject but also produces more visually pleasing and realistic images.
Besides, \ours achieves this superior perceptual quality without sacrificing geometric precision, maintaining the lowest rotation error and competitive translation error.

\begin{table*}[t]
\centering
\footnotesize

\begin{minipage}{\textwidth}
\centering
\caption{Quantitative 4D re-camera comparisons on the OpenVid dataset. BC: Background Consistency, SC: Subject Consistency, IQ: Imaging Quality, RotErr: Rotation Error, TransErr: Translation Error. Cells highlighted in \sethlcolor{red!18}\hl{red} and \sethlcolor{yellow!22}\hl{yellow} denote the best and second-best performance.}
\vspace{-3mm}

\setlength\tabcolsep{0.5mm}
\begin{tabular}{c|ccc|cc|cc}
\toprule
\hline
& \multicolumn{3}{c|}{VBench} & \multicolumn{2}{c|}{Perceptual Quality} & \multicolumn{2}{c}{Pose Accuracy} \\
\hline      
Methods & BC $\uparrow$ & SC $\uparrow$ & IQ $\uparrow$ & FVD-V $\downarrow$ & CLIP-V $\uparrow$ & RotErr $\downarrow$ & TransErr $\downarrow$\\
\hline
GEN3C & 0.9270 & 0.9067 & 0.6908 & 291.13 & 0.79 & \cellcolor{red!18}1.36 & 5.13\\
\hline
TrajCrafter &0.9235 & 0.9062 & 0.6697 & 317.08 & 0.80 & 1.38 & 5.12 \\
\hline
ReCam & 0.8977 & 0.8801 & 0.5837 & 361.98 & 0.76 & 2.15 & 5.82 \\
\hline
\hline
Scaffold-Only & 0.8898 & 0.8448 & 0.4807 & 359.38 & 0.76 & \cellcolor{yellow!22}1.37 & \cellcolor{red!18}5.10 \\
\hline
Scaffold-Early & 0.9139 & 0.9053 & 0.6172 & 273.19 & 0.83 & \cellcolor{yellow!22}1.37 & 5.13 \\
\hline
Static-Both & 0.9190 & 0.9203 & 0.6740 & \cellcolor{red!18}242.81 & \cellcolor{red!18}0.87 & 2.71 & 11.01 \\
\hline
Ours (Wan2.1) & \cellcolor{yellow!22}0.9330 & \cellcolor{red!18}0.9248 & \cellcolor{yellow!22}0.6931 & \cellcolor{yellow!22}253.13 & 0.84 & \cellcolor{yellow!22}1.37 & \cellcolor{yellow!22}5.11 \\
\hline
\hline
Ours & \cellcolor{red!18}0.9332 & \cellcolor{yellow!22}0.9247 & \cellcolor{red!18}0.6932 & 260.05 & \cellcolor{yellow!22}0.85 & \cellcolor{red!18}1.36 & 5.12 \\
\hline
\bottomrule
\end{tabular}
\label{tab:vbench}
\vspace{2mm}
\end{minipage}

\begin{minipage}{0.54\textwidth}
\centering
\includegraphics[width=\textwidth]{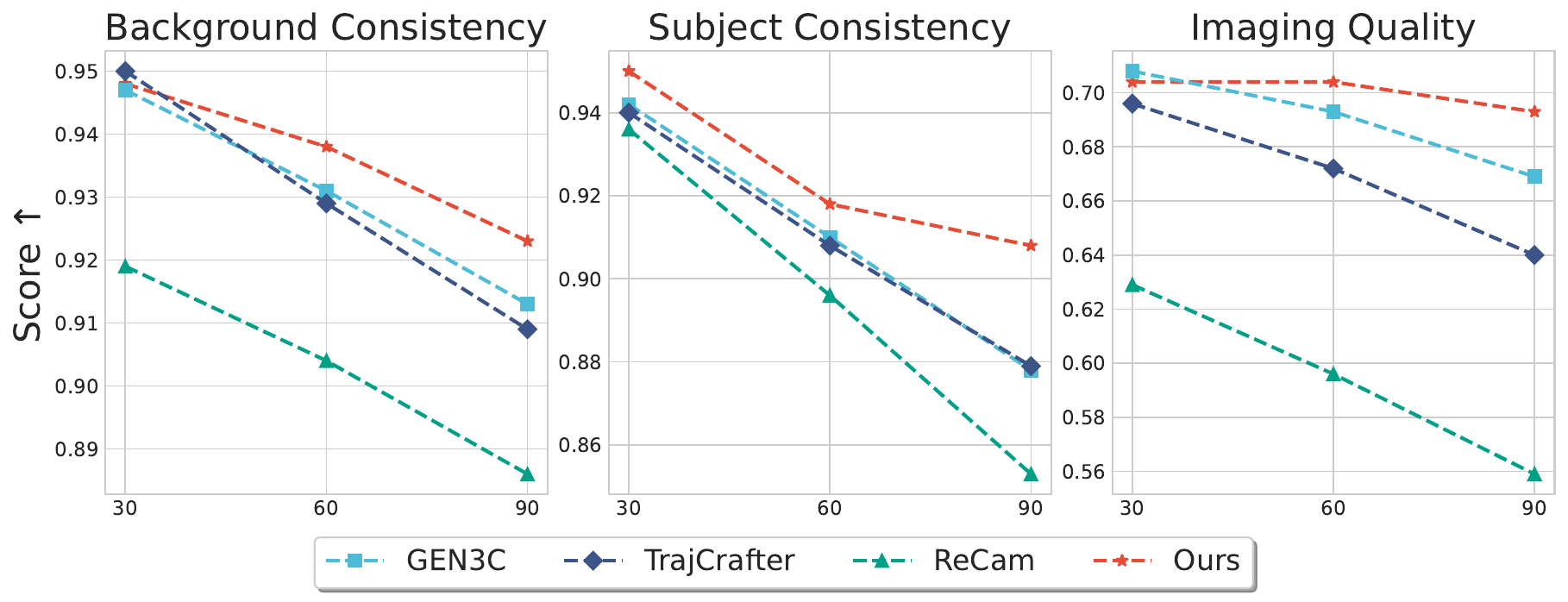}
\captionof{figure}{Quantitative results of VBench metrics on various motion magnitudes.}
\label{fig:large_quantative}
\end{minipage}
\hfill
\begin{minipage}{0.45\textwidth}
\centering
\captionof{table}{Quantitative comparisons on iPhone Dataset. Cells highlighted in red and yellow denote the best and second-best performance.}
\label{tab:multi-view}
\vspace{-3mm}
\resizebox{\textwidth}{!}{
\begin{tabular}{c|cccc}
\toprule
\hline
Methods & PSNR$\uparrow$ & SSIM$\uparrow$ & LPIPS$\downarrow$ & FVD$\downarrow$ \\
\hline
GEN3C & 12.36 & 0.4028 & 0.5112 & \cellcolor{yellow!22}260.15 \\
\hline
TrajCrafter & \cellcolor{yellow!22}13.74 & \cellcolor{yellow!22}0.4555 & \cellcolor{yellow!22}0.4819 & 273.36 \\
\hline
ReCam & 11.44 & 0.3768 & 0.5622 & 301.41 \\
\hline
Ours & \cellcolor{red!18}14.60 & \cellcolor{red!18}0.4581 & \cellcolor{red!18}0.4213 & \cellcolor{red!18}180.35 \\
\hline
\bottomrule
\end{tabular}
}
\end{minipage}
\vspace{-5mm}
\end{table*}

\begin{figure}[t]
  \centering
  \includegraphics[trim={0 0 0 0},clip,width=\linewidth]{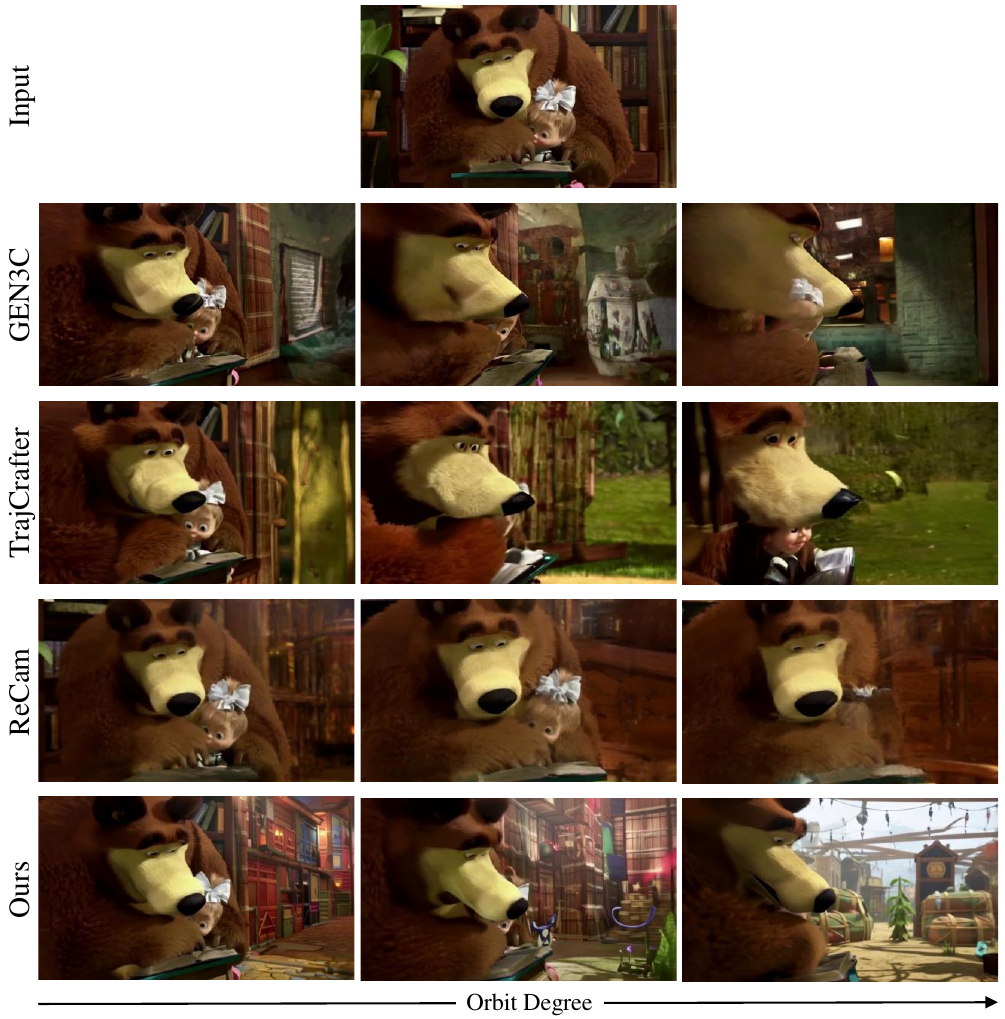}
  \vspace{-8mm}
  \caption{\footnotesize Qualitative results on various motion scales. The models are inferred under camera trajectories with three different scales of orbit degree.}
  \label{fig:large_qualitive}
  \vspace{-6mm}
\end{figure}

\parahead{Robustness to Geometric Degradation}
Geometric degradation under large view changes poses a fundamental challenge to all view synthesis paradigms. For scaffold-based methods, large camera motions induce severe geometric sparsity (disocclusion holes, depth inaccuracy); for scaffold-free implicit methods like ReCamMaster, the same motions cause geometric drift due to the lack of explicit spatial constraints. We design an experiment to measure robustness to this unified challenge, using motion magnitude as a controlled proxy to induce progressive geometric degradation.
Starting from a modest 30-degree orbit (minimal geometric stress), we progressively increase motion magnitude to a challenging 90-degree trajectory. At 90 degrees, scaffold-based methods face extremely sparse geometry, while implicit methods face severe misalignment between source and target views. Fig.~\ref{fig:large_quantative} plots performance as geometric degradation intensifies with increasing motion.
All competitors deteriorate under this stress test: GEN3C and TrajectoryCrafter propagate errors from sparse, hole-ridden point clouds; ReCamMaster, despite being scaffold-free, suffers catastrophic geometric drift without early-stage structural anchoring. \ours maintains consistently high scores because its stage-wise decoupling addresses both failure modes: the early geometry-anchoring stage prevents drift (solving ReCamMaster's vulnerability), while the late appearance stage corrects sparse geometric errors (solving scaffold-based methods' vulnerability).
The qualitative results in Fig.~\ref{fig:large_qualitive} confirm this: as geometric degradation increases, previous methods produce warped, broken, or drifted figures, while \ours renders coherent subjects by leveraging geometry for structure without being corrupted by its sparsity.

\subsection{Evaluation on Multi-view Video Benchmark}
While our primary focus is on in-the-wild monocular videos, we also conduct experiments on a multi-view dataset to enable evaluation with pixel-wise metrics. Following the setup of TrajectoryCrafter~\cite{yu2025trajectorycrafter}, we use the iPhone dataset~\cite{gao2022dynamic}, treating one moving camera view as the monocular input and a static camera view as the ground-truth target.

\begin{figure}[t]
  \centering
  \includegraphics[trim={0 0 0 0},clip,width=\linewidth]{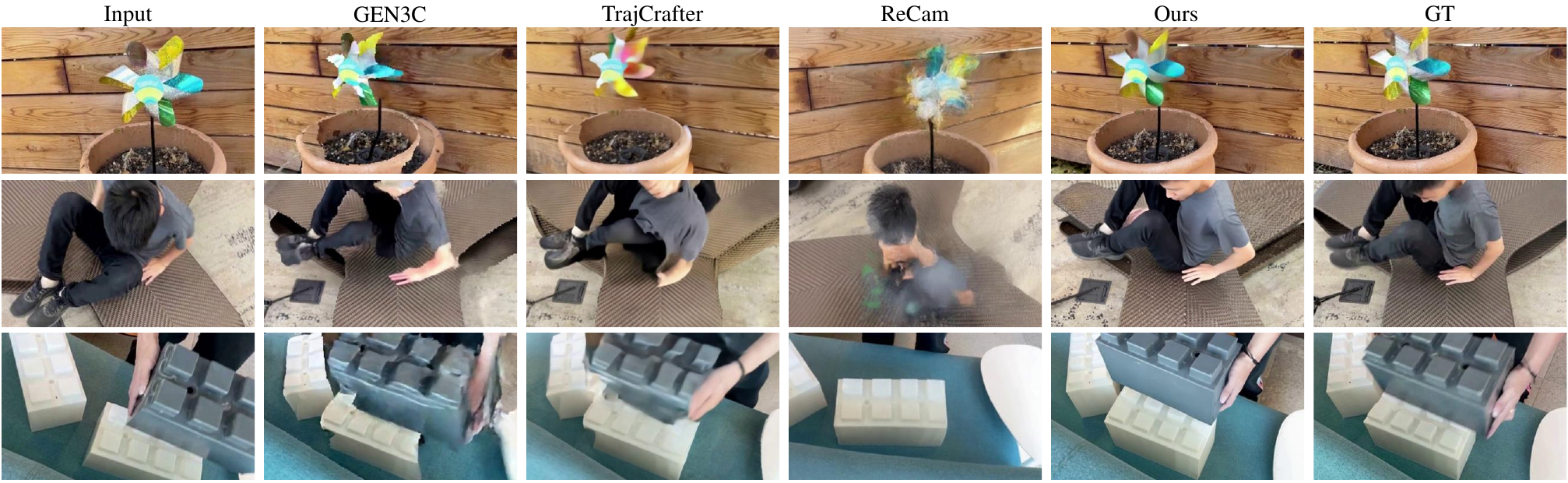}
  \vspace{-8mm}
  \caption{\footnotesize Qualitative results on iPhone Dataset.}
  \vspace{-3mm}
  \label{fig:multi-view}
\end{figure}

Tab.~\ref{tab:multi-view} shows that \ours significantly outperforms other methods on PSNR, SSIM, LPIPS and FVD. The strong improvement in LPIPS and FVD, perceptual metrics for perceptual quality, is particularly noteworthy, indicating that our generated views are perceptually closer to the ground truth. This demonstrates that our timestep-gated conditioning not only improves general coherence but also preserves geometric and appearance details with high fidelity. The qualitative results in Fig.~\ref{fig:multi-view} corroborate this; for instance, notice the finer details on the subject's clothing and the more accurate facial structure rendered by our method compared to the blurry or distorted results from others.

\subsection{Ablation Studies}
\label{sec:ablation}
We conduct a series of ablation studies to dissect the contributions of our key design choices.

\begin{figure}[t]
  \centering
  \includegraphics[trim={0 0 0 0},clip,width=\linewidth]{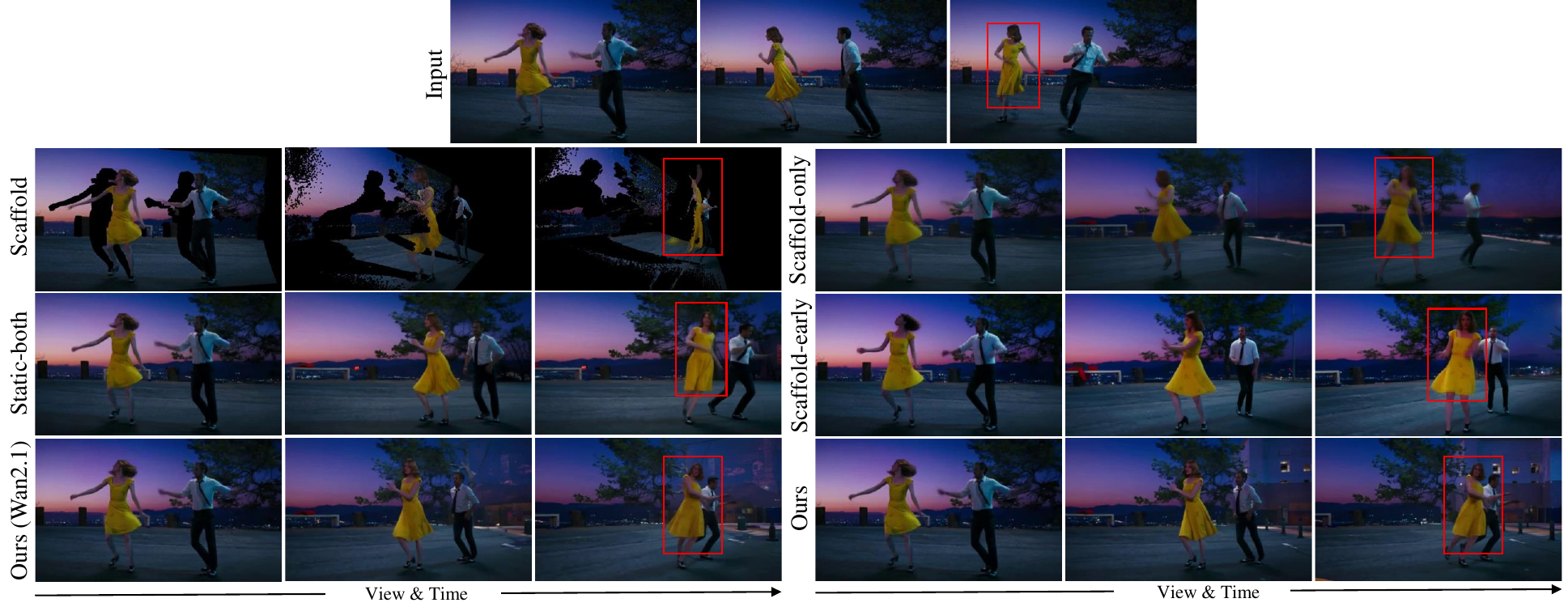}
  \vspace{-8mm}
  \caption{\footnotesize Ablation results on structured denoising dynamics.}
  \vspace{-6mm}
  \label{fig:ablation_guidance}
\end{figure}

\parahead{Effectiveness of Structured Denoising Dynamics}
To validate our core hypothesis that a temporally-aligned structured denoising generation is essential, we compare \ours against several variants.
The variants are: (1) \textit{Scaffold-Only}: The model is conditioned only on the scaffold video $c^{ren}$ for all timesteps. (2) \textit{Scaffold-Early}: The model is conditioned on the scaffold video $c^{ren}$ only during the early timesteps ($t > T_{switch}$), with no explicit conditioning in the later stages. This variant tests the hypothesis that simply removing the flawed scaffold signal is sufficient to mitigate artifact propagation. (3) \textit{Static-Both}: Both scaffold and reference conditions (\ie, $c^{ren}$ and $c^{src}$) are provided simultaneously throughout the entire denoising process.

As shown in Tab.~\ref{tab:vbench}, \ours significantly outperforms all variants, demonstrating the advantage of the proposed structured denoising dynamics, and Fig.~\ref{fig:ablation_guidance} qualitatively illustrate the results.
The `Scaffold-Only' baseline collapses across all metrics (IQ merely 0.4807), proving that persistent geometric conditioning permanently bakes scaffold errors into the output.
The `Scaffold-Early' model successfully avoids inheriting the worst point cloud artifacts, but without the reference signal in the later stages, it struggles to synthesize fine-grained, scene-consistent textures and often produces blurry or generic details in disoccluded regions. 
The `Static-Both' model suffers catastrophic geometric instability (Rotation Error 2.71, Translation Error 11.01) despite competitive perceptual scores, validating that simultaneous conditioning creates signal interference.
In contrast, \ours leverages the strengths of both signals in sequence: it first establishes correct geometry with the scaffold and then actively corrects its flaws and refines photorealism using the reference video. These results validate that merely removing imperfect geometry (Scaffold-Early) or adding appearance statically (Static-Both) is insufficient; the deliberate stage-wise handover is essential to resolve the geometry-appearance conflict.

\parahead{Generalization across Different Backbones}
To further validate the robustness of our approach, we extend our ablation studies to different video generation backbones. Specifically, we further evaluate our method on both Wan2.1. As illustrated in Fig.~\ref{fig:ablation_guidance}, the qualitative comparison demonstrates that our method achieves comparable high-fidelity results on Wan2.1, similar to the performance observed on Wan2.2. This consistency across different architectures confirms that our method is robust to the choice of backbone. More importantly, it provides strong evidence that the performance gains stem from our proposed structured denoising dynamics, rather than depending on a specific video model architecture. Quantitative results in Tab.~\ref{tab:vbench} also demonstrate its generalization ability.

\parahead{Robustness of Depth Estimation}
Monocular depth estimation inevitably introduces inaccuracies that propagate into point cloud reconstructions. Our stage-wise mechanism tolerates these imperfections: early geometry anchoring prevents structural drift, while late appearance correction rectifies artifacts 
\begin{wrapfigure}{r}{5cm}
  \vspace{-5mm}
  \centering
  \includegraphics[width=5cm]{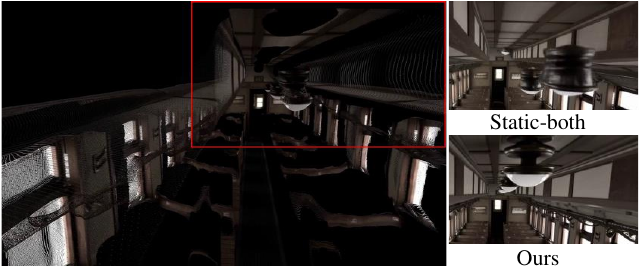}
  \vspace{-7mm}
  \caption{\footnotesize Depth Robustness.}
  \vspace{-5mm}
  \label{fig:depth}
\end{wrapfigure}
before they are baked into the final output. Fig.~\ref{fig:depth} validates this robustness under depth errors, where \ours maintains coherence despite perturbed inputs: lights distorted by inaccurate depth in the scaffold are successfully corrected in the final output, demonstrating that \ours explicitly compensates for reconstruction noise rather than relying on perfect geometry.

\section{Conclusion}
\label{sec:conclusion}
We presented \ours, a unified 3D/4D view synthesis framework that addresses sparse and erroneous geometry through structured denoising dynamics. By temporally decoupling geometry and appearance, early scaffold anchoring followed by late-stage error correction, our method prevents propagation of point cloud flaws without sacrificing geometric control. This stage-wise design achieves robustness to the imperfect reconstruction inevitable in monocular settings. Future work may explore joint scaffold-video refinement.


%
%
\bibliographystyle{splncs04}
\bibliography{main}

@String(CVPR= {IEEE Conf. Comput. Vis. Pattern Recog.})

@String(ICCV= {Int. Conf. Comput. Vis.})

@String(AAAI = {AAAI})

@String(CVPR  = {CVPR})

@String(ICCV  = {ICCV})

@article{bai2025recammaster,
  title={Recammaster: Camera-controlled generative rendering from a single video},
  author={Bai, Jianhong and Xia, Menghan and Fu, Xiao and Wang, Xintao and Mu, Lianrui and Cao, Jinwen and Liu, Zuozhu and Hu, Haoji and Bai, Xiang and Wan, Pengfei and others},
  journal={arXiv preprint arXiv:2503.11647},
  year={2025}
}

@article{yu2025trajectorycrafter,
  title={Trajectorycrafter: Redirecting camera trajectory for monocular videos via diffusion models},
  author={YU, Mark and Hu, Wenbo and Xing, Jinbo and Shan, Ying},
  journal={arXiv preprint arXiv:2503.05638},
  year={2025}
}

@inproceedings{van2024generative,
  title={Generative camera dolly: Extreme monocular dynamic novel view synthesis},
  author={Van Hoorick, Basile and Wu, Rundi and Ozguroglu, Ege and Sargent, Kyle and Liu, Ruoshi and Tokmakov, Pavel and Dave, Achal and Zheng, Changxi and Vondrick, Carl},
  booktitle={European Conference on Computer Vision},
  pages={313--331},
  year={2024},
  organization={Springer}
}

@inproceedings{ren2025gen3c,
  title={Gen3c: 3d-informed world-consistent video generation with precise camera control},
  author={Ren, Xuanchi and Shen, Tianchang and Huang, Jiahui and Ling, Huan and Lu, Yifan and Nimier-David, Merlin and M{\"u}ller, Thomas and Keller, Alexander and Fidler, Sanja and Gao, Jun},
  booktitle={Proceedings of the Computer Vision and Pattern Recognition Conference},
  pages={6121--6132},
  year={2025}
}

@inproceedings{chen2025cognvs,
  title     = {Reconstruct, Inpaint, Finetune: Dynamic Novel-view Synthesis from Monocular Videos},
  author    = {Chen, Kaihua and Khurana, Tarasha and Ramanan, Deva},
  booktitle = {Advances in Neural Information Processing Systems (NeurIPS)},
  year      = {2025}
}

@article{blattmann2023stable,
  title={Stable video diffusion: Scaling latent video diffusion models to large datasets},
  author={Blattmann, Andreas and Dockhorn, Tim and Kulal, Sumith and Mendelevitch, Daniel and Kilian, Maciej and Lorenz, Dominik and Levi, Yam and English, Zion and Voleti, Vikram and Letts, Adam and others},
  journal={arXiv preprint arXiv:2311.15127},
  year={2023}
}

@article{wan2025,
      title={Wan: Open and Advanced Large-Scale Video Generative Models}, 
      author={Team Wan and Ang Wang and Baole Ai and Bin Wen and Chaojie Mao and Chen-Wei Xie and Di Chen and Feiwu Yu and Haiming Zhao and Jianxiao Yang and Jianyuan Zeng and Jiayu Wang and Jingfeng Zhang and Jingren Zhou and Jinkai Wang and Jixuan Chen and Kai Zhu and Kang Zhao and Keyu Yan and Lianghua Huang and Mengyang Feng and Ningyi Zhang and Pandeng Li and Pingyu Wu and Ruihang Chu and Ruili Feng and Shiwei Zhang and Siyang Sun and Tao Fang and Tianxing Wang and Tianyi Gui and Tingyu Weng and Tong Shen and Wei Lin and Wei Wang and Wei Wang and Wenmeng Zhou and Wente Wang and Wenting Shen and Wenyuan Yu and Xianzhong Shi and Xiaoming Huang and Xin Xu and Yan Kou and Yangyu Lv and Yifei Li and Yijing Liu and Yiming Wang and Yingya Zhang and Yitong Huang and Yong Li and You Wu and Yu Liu and Yulin Pan and Yun Zheng and Yuntao Hong and Yupeng Shi and Yutong Feng and Zeyinzi Jiang and Zhen Han and Zhi-Fan Wu and Ziyu Liu},
      journal = {arXiv preprint arXiv:2503.20314},
      year={2025}
}

@article{yang2024cogvideox,
  title={Cogvideox: Text-to-video diffusion models with an expert transformer},
  author={Yang, Zhuoyi and Teng, Jiayan and Zheng, Wendi and Ding, Ming and Huang, Shiyu and Xu, Jiazheng and Yang, Yuanming and Hong, Wenyi and Zhang, Xiaohan and Feng, Guanyu and others},
  journal={arXiv preprint arXiv:2408.06072},
  year={2024}
}

@article{kong2024hunyuanvideo,
  title={Hunyuanvideo: A systematic framework for large video generative models},
  author={Kong, Weijie and Tian, Qi and Zhang, Zijian and Min, Rox and Dai, Zuozhuo and Zhou, Jin and Xiong, Jiangfeng and Li, Xin and Wu, Bo and Zhang, Jianwei and others},
  journal={arXiv preprint arXiv:2412.03603},
  year={2024}
}

@inproceedings{gao2021dynamic,
  title={Dynamic view synthesis from dynamic monocular video},
  author={Gao, Chen and Saraf, Ayush and Kopf, Johannes and Huang, Jia-Bin},
  booktitle={Proceedings of the IEEE/CVF International Conference on Computer Vision},
  pages={5712--5721},
  year={2021}
}

@inproceedings{wu2025cat4d,
  title={Cat4d: Create anything in 4d with multi-view video diffusion models},
  author={Wu, Rundi and Gao, Ruiqi and Poole, Ben and Trevithick, Alex and Zheng, Changxi and Barron, Jonathan T and Holynski, Aleksander},
  booktitle={Proceedings of the Computer Vision and Pattern Recognition Conference},
  pages={26057--26068},
  year={2025}
}

@inproceedings{mildenhall2020nerf,
  title={NeRF: Representing Scenes as Neural Radiance Fields for View Synthesis},
  author={Mildenhall, Ben and Srinivasan, Pratul P and Tancik, Matthew and Barron, Jonathan T and Ramamoorthi, Ravi and Ng, Ren},
  booktitle={European Conference on Computer Vision},
  pages={405--421},
  year={2020},
  organization={Springer}
}

@article{kerbl20233d,
  title={3D Gaussian splatting for real-time radiance field rendering.},
  author={Kerbl, Bernhard and Kopanas, Georgios and Leimk{\"u}hler, Thomas and Drettakis, George},
  journal={ACM Trans. Graph.},
  volume={42},
  number={4},
  pages={139--1},
  year={2023}
}

@inproceedings{li2021neural,
  title={Neural scene flow fields for space-time view synthesis of dynamic scenes},
  author={Li, Zhengqi and Niklaus, Simon and Snavely, Noah and Wang, Oliver},
  booktitle={Proceedings of the IEEE/CVF conference on computer vision and pattern recognition},
  pages={6498--6508},
  year={2021}
}

@inproceedings{fridovich2023k,
  title={K-planes: Explicit radiance fields in space, time, and appearance},
  author={Fridovich-Keil, Sara and Meanti, Giacomo and Warburg, Frederik Rahb{\ae}k and Recht, Benjamin and Kanazawa, Angjoo},
  booktitle={Proceedings of the IEEE/CVF Conference on Computer Vision and Pattern Recognition},
  pages={12479--12488},
  year={2023}
}

@inproceedings{cao2023hexplane,
  title={Hexplane: A fast representation for dynamic scenes},
  author={Cao, Ang and Johnson, Justin},
  booktitle={Proceedings of the IEEE/CVF Conference on Computer Vision and Pattern Recognition},
  pages={130--141},
  year={2023}
}

@inproceedings{li2022neural,
  title={Neural 3d video synthesis from multi-view video},
  author={Li, Tianye and Slavcheva, Mira and Zollhoefer, Michael and Green, Simon and Lassner, Christoph and Kim, Changil and Schmidt, Tanner and Lovegrove, Steven and Goesele, Michael and Newcombe, Richard and others},
  booktitle={Proceedings of the IEEE/CVF conference on computer vision and pattern recognition},
  pages={5521--5531},
  year={2022}
}

@article{yang2023real,
  title={Real-time photorealistic dynamic scene representation and rendering with 4d gaussian splatting},
  author={Yang, Zeyu and Yang, Hongye and Pan, Zijie and Zhang, Li},
  journal={arXiv preprint arXiv:2310.10642},
  year={2023}
}

@inproceedings{yang2024deformable,
  title={Deformable 3d gaussians for high-fidelity monocular dynamic scene reconstruction},
  author={Yang, Ziyi and Gao, Xinyu and Zhou, Wen and Jiao, Shaohui and Zhang, Yuqing and Jin, Xiaogang},
  booktitle={Proceedings of the IEEE/CVF conference on computer vision and pattern recognition},
  pages={20331--20341},
  year={2024}
}

@inproceedings{wu20244d,
  title={4d gaussian splatting for real-time dynamic scene rendering},
  author={Wu, Guanjun and Yi, Taoran and Fang, Jiemin and Xie, Lingxi and Zhang, Xiaopeng and Wei, Wei and Liu, Wenyu and Tian, Qi and Wang, Xinggang},
  booktitle={Proceedings of the IEEE/CVF conference on computer vision and pattern recognition},
  pages={20310--20320},
  year={2024}
}

@inproceedings{lin2024gaussian,
  title={Gaussian-flow: 4d reconstruction with dynamic 3d gaussian particle},
  author={Lin, Youtian and Dai, Zuozhuo and Zhu, Siyu and Yao, Yao},
  booktitle={Proceedings of the IEEE/CVF Conference on Computer Vision and Pattern Recognition},
  pages={21136--21145},
  year={2024}
}

@inproceedings{li2024spacetime,
  title={Spacetime gaussian feature splatting for real-time dynamic view synthesis},
  author={Li, Zhan and Chen, Zhang and Li, Zhong and Xu, Yi},
  booktitle={Proceedings of the IEEE/CVF Conference on Computer Vision and Pattern Recognition},
  pages={8508--8520},
  year={2024}
}

@inproceedings{duan20244d,
  title={4d-rotor gaussian splatting: towards efficient novel view synthesis for dynamic scenes},
  author={Duan, Yuanxing and Wei, Fangyin and Dai, Qiyu and He, Yuhang and Chen, Wenzheng and Chen, Baoquan},
  booktitle={ACM SIGGRAPH 2024 Conference Papers},
  pages={1--11},
  year={2024}
}

@article{zhu2025dynamic,
  title={Dynamic scene reconstruction: Recent advance in real-time rendering and streaming},
  author={Zhu, Jiaxuan and Tang, Hao},
  journal={arXiv preprint arXiv:2503.08166},
  year={2025}
}

@inproceedings{pumarola2020d,
  title={D-nerf: Neural radiance fields for dynamic scenes. 2021 IEEE},
  author={Pumarola, Albert and Corona, Enric and Pons-Moll, Gerard and Moreno-Noguer, Francesc},
  booktitle={CVF Conference on Computer Vision and Pattern Recognition (CVPR)},
  pages={10313--10322},
  year={2020}
}

@article{luo2025instant4d,
  title={Instant4d: 4d gaussian splatting in minutes},
  author={Luo, Zhanpeng and Ran, Haoxi and Lu, Li},
  journal={Advances in neural information processing systems},
  year={2025}
}

@inproceedings{
    liu2025modgs,
    title={Mo{DGS}: Dynamic Gaussian Splatting from Casually-captured Monocular Videos with Depth Priors},
    author={Qingming LIU and Yuan Liu and Jiepeng Wang and Xianqiang Lyu and Peng Wang and Wenping Wang and Junhui Hou},
    booktitle={The Thirteenth International Conference on Learning Representations},
    year={2025},
    url={https://openreview.net/forum?id=2prShxdLkX}
}

@InProceedings{Fan_2025_CVPR,
    author    = {Fan, Cheng-De and Chang, Chen-Wei and Liu, Yi-Ruei and Lee, Jie-Ying and Huang, Jiun-Long and Tseng, Yu-Chee and Liu, Yu-Lun},
    title     = {SpectroMotion: Dynamic 3D Reconstruction of Specular Scenes},
    booktitle = {Proceedings of the IEEE/CVF Conference on Computer Vision and Pattern Recognition (CVPR)},
    month     = {June},
    year      = {2025},
    pages     = {21328-21338}
}

@InProceedings{Zhang_2025_ICCV,
    author    = {Zhang, Xinjie and Liu, Zhening and Zhang, Yifan and Ge, Xingtong and He, Dailan and Xu, Tongda and Wang, Yan and Lin, Zehong and Yan, Shuicheng and Zhang, Jun},
    title     = {MEGA: Memory-Efficient 4D Gaussian Splatting for Dynamic Scenes},
    booktitle = {Proceedings of the IEEE/CVF International Conference on Computer Vision (ICCV)},
    month     = {October},
    year      = {2025},
    pages     = {27828-27838}
}

@InProceedings{Song_2025_ICCV,
    author    = {Song, Rui and Liang, Chenwei and Xia, Yan and Zimmer, Walter and Cao, Hu and Caesar, Holger and Festag, Andreas and Knoll, Alois},
    title     = {CoDa-4DGS: Dynamic Gaussian Splatting with Context and Deformation Awareness for Autonomous Driving},
    booktitle = {Proceedings of the IEEE/CVF International Conference on Computer Vision (ICCV)},
    month     = {October},
    year      = {2025},
    pages     = {28031-28041}
}

@article{you2024nvs,
  title={Nvs-solver: Video diffusion model as zero-shot novel view synthesizer},
  author={You, Meng and Zhu, Zhiyu and Liu, Hui and Hou, Junhui},
  journal={arXiv preprint arXiv:2405.15364},
  year={2024}
}

@inproceedings{zhang2025recapture,
  title={Recapture: Generative video camera controls for user-provided videos using masked video fine-tuning},
  author={Zhang, David Junhao and Paiss, Roni and Zada, Shiran and Karnad, Nikhil and Jacobs, David E and Pritch, Yael and Mosseri, Inbar and Shou, Mike Zheng and Wadhwa, Neal and Ruiz, Nataniel},
  booktitle={Proceedings of the Computer Vision and Pattern Recognition Conference},
  pages={2050--2062},
  year={2025}
}

@article{jeong2025reangle,
  title={Reangle-a-video: 4d video generation as video-to-video translation},
  author={Jeong, Hyeonho and Lee, Suhyeon and Ye, Jong Chul},
  journal={arXiv preprint arXiv:2503.09151},
  year={2025}
}

@article{lei2025motionflow,
  title={MotionFlow: Learning Implicit Motion Flow for Complex Camera Trajectory Control in Video Generation},
  author={Lei, Guojun and Wang, Chi and Wang, Yikai and Li, Hong and Song, Ying and Xu, Weiwei},
  journal={arXiv preprint arXiv:2509.21119},
  year={2025}
}

@article{bahmani2024vd3d,
  title={Vd3d: Taming large video diffusion transformers for 3d camera control},
  author={Bahmani, Sherwin and Skorokhodov, Ivan and Siarohin, Aliaksandr and Menapace, Willi and Qian, Guocheng and Vasilkovsky, Michael and Lee, Hsin-Ying and Wang, Chaoyang and Zou, Jiaxu and Tagliasacchi, Andrea and others},
  journal={arXiv preprint arXiv:2407.12781},
  year={2024}
}

@misc{zhang2023adding,
  title={Adding Conditional Control to Text-to-Image Diffusion Models}, 
  author={Lvmin Zhang and Anyi Rao and Maneesh Agrawala},
  booktitle={IEEE International Conference on Computer Vision (ICCV)},
  year={2023},
}

@inproceedings{mou2024t2i,
  title={T2i-adapter: Learning adapters to dig out more controllable ability for text-to-image diffusion models},
  author={Mou, Chong and Wang, Xintao and Xie, Liangbin and Wu, Yanze and Zhang, Jian and Qi, Zhongang and Shan, Ying},
  booktitle={Proceedings of the AAAI conference on artificial intelligence},
  volume={38},
  number={5},
  pages={4296--4304},
  year={2024}
}

@article{ye2023ip,
  title={Ip-adapter: Text compatible image prompt adapter for text-to-image diffusion models},
  author={Ye, Hu and Zhang, Jun and Liu, Sibo and Han, Xiao and Yang, Wei},
  journal={arXiv preprint arXiv:2308.06721},
  year={2023}
}

@inproceedings{ham2025diffusion,
  title={Diffusion Model Patching via Mixture-of-Prompts},
  author={Ham, Seokil and Woo, Sangmin and Kim, Jin-Young and Go, Hyojun and Park, Byeongjun and Kim, Changick},
  booktitle={Proceedings of the AAAI Conference on Artificial Intelligence},
  volume={39},
  number={16},
  pages={17023--17031},
  year={2025}
}

@article{zhuang2025timestep,
  title={TimeStep Master: Asymmetrical Mixture of Timestep LoRA Experts for Versatile and Efficient Diffusion Models in Vision},
  author={Zhuang, Shaobin and Guo, Yiwei and Ding, Yanbo and Li, Kunchang and Chen, Xinyuan and Wang, Yaohui and Wang, Fangyikang and Zhang, Ying and Li, Chen and Wang, Yali},
  journal={arXiv preprint arXiv:2503.07416},
  year={2025}
}

@inproceedings{wu2025improved,
  title={Improved video vae for latent video diffusion model},
  author={Wu, Pingyu and Zhu, Kai and Liu, Yu and Zhao, Liming and Zhai, Wei and Cao, Yang and Zha, Zheng-Jun},
  booktitle={Proceedings of the Computer Vision and Pattern Recognition Conference},
  pages={18124--18133},
  year={2025}
}

@inproceedings{gao2022dynamic,
    title={Dynamic Novel-View Synthesis: A Reality Check},
    author={Gao, Hang and Li, Ruilong and Tulsiani, Shubham and Russell, Bryan and Kanazawa, Angjoo},
    booktitle={NeurIPS},
    year={2022},
}

@article{nan2024openvid,
  title={OpenVid-1M: A Large-Scale High-Quality Dataset for Text-to-video Generation},
  author={Nan, Kepan and Xie, Rui and Zhou, Penghao and Fan, Tiehan and Yang, Zhenheng and Chen, Zhijie and Li, Xiang and Yang, Jian and Tai, Ying},
  journal={arXiv preprint arXiv:2407.02371},
  year={2024}
}

@InProceedings{huang2023vbench,
 title={{VBench}: Comprehensive Benchmark Suite for Video Generative Models},
 author={Huang, Ziqi and He, Yinan and Yu, Jiashuo and Zhang, Fan and Si, Chenyang and Jiang, Yuming and Zhang, Yuanhan and Wu, Tianxing and Jin, Qingyang and Chanpaisit, Nattapol and Wang, Yaohui and Chen, Xinyuan and Wang, Limin and Lin, Dahua and Qiao, Yu and Liu, Ziwei},
 booktitle={Proceedings of the IEEE/CVF Conference on Computer Vision and Pattern Recognition},
 year={2024}
}

@article{huang2025vipe,
  title={Vipe: Video pose engine for 3d geometric perception},
  author={Huang, Jiahui and Zhou, Qunjie and Rabeti, Hesam and Korovko, Aleksandr and Ling, Huan and Ren, Xuanchi and Shen, Tianchang and Gao, Jun and Slepichev, Dmitry and Lin, Chen-Hsuan and others},
  journal={arXiv preprint arXiv:2508.10934},
  year={2025}
}

@article{yu2025viewcrafter,
  title={ViewCrafter: Taming Video Diffusion Models for High-fidelity Novel View Synthesis},
  author={Yu, Wangbo and Xing, Jinbo and Yuan, Li and Hu, Wenbo and Li, Xiaoyu and Huang, Zhipeng and Gao, Xiangjun and Wong, Tien-Tsin and Shan, Ying and Tian, Yonghong},
  journal={IEEE Transactions on Pattern Analysis and Machine Intelligence},
  year={2025},
  publisher={IEEE}
}

@inproceedings{wang2025vistadream,
  title={Vistadream: Sampling multiview consistent images for single-view scene reconstruction},
  author={Wang, Haiping and Liu, Yuan and Liu, Ziwei and Wang, Wenping and Dong, Zhen and Yang, Bisheng},
  booktitle={Proceedings of the IEEE/CVF International Conference on Computer Vision},
  pages={26772--26782},
  year={2025}
}

@inproceedings{zhang2025spatialcrafter,
  title={Spatialcrafter: Unleashing the imagination of video diffusion models for scene reconstruction from limited observations},
  author={Zhang, Songchun and Xu, Huiyao and Guo, Sitong and Xie, Zhongwei and Bao, Hujun and Xu, Weiwei and Zou, Changqing},
  booktitle={Proceedings of the IEEE/CVF International Conference on Computer Vision},
  pages={27794--27805},
  year={2025}
}

@inproceedings{shriram2025realmdreamer,
  title={RealmDreamer: Text-Driven 3D Scene Generation with Inpainting and Depth Diffusion},
  author={Shriram, Jaidev and Trevithick, Alex and Liu, Lingjie and Ramamoorthi, Ravi},
  booktitle={2025 International Conference on 3D Vision (3DV)},
  pages={882--892},
  year={2025},
  organization={IEEE}
}

@article{zhang2024text2nerf,
  title={Text2nerf: Text-driven 3d scene generation with neural radiance fields},
  author={Zhang, Jingbo and Li, Xiaoyu and Wan, Ziyu and Wang, Can and Liao, Jing},
  journal={IEEE Transactions on Visualization and Computer Graphics},
  volume={30},
  number={12},
  pages={7749--7762},
  year={2024},
  publisher={IEEE}
}

@article{chung2025luciddreamer,
  title={LucidDreamer: Domain-free Generation of 3D Gaussian Splatting Scenes},
  author={Chung, Jaeyoung and Lee, Suyoung and Nam, Hyeongjin and Lee, Jaerin and Lee, Kyoung Mu},
  journal={IEEE Transactions on Visualization and Computer Graphics},
  year={2025},
  publisher={IEEE}
}

@article{he2024cameractrl,
  title={Cameractrl: Enabling camera control for text-to-video generation},
  author={He, Hao and Xu, Yinghao and Guo, Yuwei and Wetzstein, Gordon and Dai, Bo and Li, Hongsheng and Yang, Ceyuan},
  journal={arXiv preprint arXiv:2404.02101},
  year={2024}
}

\clearpage
\appendix
\crefalias{section}{appendix}
\crefalias{subsection}{appendix}

\section{Comparison with 3D-based method}
\begin{figure}[h]
  \centering
  \includegraphics[trim={0 0 0 0},clip,width=\linewidth]{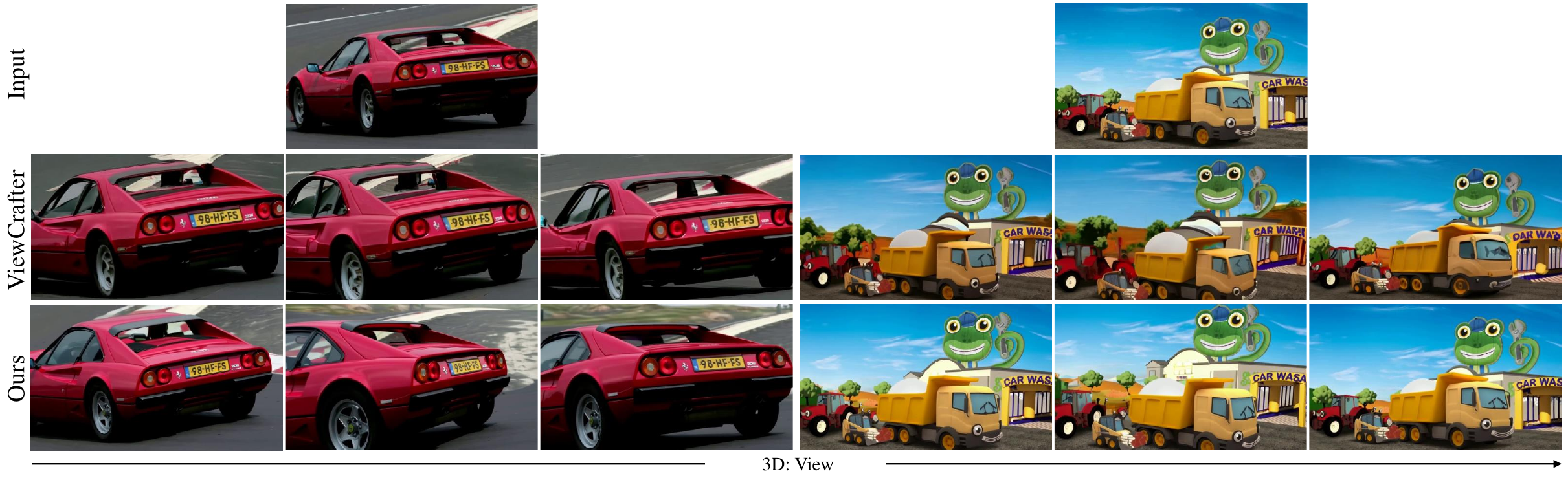}
  \vspace{-8mm}
  \caption{\footnotesize Qualitative results for single-view 3D reconstruction.}
  \vspace{-8mm}
  \label{fig:supp_qualitative_3d}
\end{figure}

We further compare our method with ViewCrafter~\cite{yu2025viewcrafter} on single-view 3D reconstruction both qualitatively and quantitatively. Fig.~\ref{fig:supp_qualitative_3d} shows that ViewCrafter introduce geometry distortion and view coherence (\eg, the tire in the first sample and the text in the second sample), while ours maintain correct shape due to its structured denoising dynamic mechanism. The quantitative results in Tab.~\ref{tab:supp_quantitative_3d} also demonstrates our method achieve better visual quality and structure accuracy as well.

\begin{table}[h]
	\caption{Quantitative 3D reconstruction comparisons on the OpenVid dataset. BC: Background Consistency, SC: Subject Consistency, IQ: Imaging Quality, RotErr: Rotation Error, TransErr: Translation Error.}
    \vspace{-3mm}
	\label{tab:supp_quantitative_3d}
    \centering
    \footnotesize
    \setlength\tabcolsep{0.5mm}
    \begin{tabular}{c|ccc|cc|cc}
        \toprule
        \hline
        & \multicolumn{3}{c|}{VBench} & \multicolumn{2}{c|}{Perceptual Quality} & \multicolumn{2}{c}{Pose Accuracy} \\
		\hline      
		Methods & BC $\uparrow$ & SC $\uparrow$ & IQ $\uparrow$ & FVD-V $\downarrow$ & CLIP-V $\uparrow$ & RotErr $\downarrow$ & TransErr $\downarrow$\\
        \hline
		ViewCrafter & 0.9268 & 0.9082 & 0.6836 &  308.24 &  0.79 & 1.39 & 5.14\\
        \hline
        Ours  & \textbf{0.9334} & \textbf{0.9250} & \textbf{0.6961} &  \textbf{255.16} &  \textbf{0.87} &  \textbf{1.35} &  \textbf{5.11} \\
        \hline
    \bottomrule
	\end{tabular}
    \vspace{-4mm}
\end{table}

\end{document}